\documentclass[letterpaper, 10 pt, conference]{ieeeconf}  

\IEEEoverridecommandlockouts                              

\usepackage{graphics} 
\usepackage{caption}
\usepackage{subcaption}
\usepackage{amsmath} 
\usepackage{amssymb}  
\usepackage{mathtools} 
\usepackage{algorithm,algcompatible}
\usepackage{tabularray}
\UseTblrLibrary{siunitx}

\graphicspath{ {images/} }

\newcommand{\R}{\mathbb{R}}

\newcommand{\z}{\mathbf{z}}

\newcommand{\uu}{\mathbf{u}}
\newcommand{\s}{\mathbf{s}}

\newcommand{\Ss}{\mathcal{S}}

\newcommand{\X}{\mathcal{X}}
\newcommand{\U}{\mathcal{U}}

\newcommand{\I}{\mathcal{I}}

\newcommand{\Rr}{\mathcal{R}}
\newcommand{\Pp}{\mathcal{P}}

\newtheorem{definition}{Definition}

\title{\LARGE \bf
Training Adversarial yet Safe Agent to Characterize Safety Performance of Highly Automated Vehicles
}

\author{Minghao Zhu$^{1}$$^{2}$, Anmol Sidhu$^{1}$ and Keith A. Redmill$^{2}$
\thanks{$^{1}$Minghao Zhu and Anmol Sidhu are with the Advanced Mobility, Transportation Research Center Inc.,
        East Liberty, OH 43319, USA
        {\tt\small }}%
\thanks{$^{2}$Minghao Zhu and Keith A. Redmill are with the Department of Electrical and Computer Engineering, The Ohio State University,
        Columbus, OH 43210, USA
        {\tt\small redmill.1@osu.edu}}%
}

\begin{document}

\maketitle
\thispagestyle{empty}
\pagestyle{empty}

\begin{abstract}
This paper focuses on safety performance testing and characterization of black-box highly automated vehicles (HAV). Existing testing approaches typically obtain the testing outcomes by deploying the HAV into a specific testing environment. Such a testing environment can involve various passively given testing strategies presented by other traffic participants such as (i) the naturalistic driving policy learned from human drivers, (ii) extracted concrete scenarios from real-world driving data, and (iii) model-based or data-driven adversarial testing methodologies focusing on forcing safety-critical events. The safety performance of HAV is further characterized by analyzing the obtained testing outcomes with a particular selected measure, such as the observed collision risk. The aforementioned testing practices suffer from the scarcity of safety-critical events, have limited operational design domain (ODD) coverage, or are biased toward long-tail unsafe cases. This paper presents a novel and informative testing strategy that differs from these existing practices. The proposal is inspired by the intuition that a relatively safer HAV driving policy would allow the traffic vehicles to exhibit a higher level of aggressiveness to achieve a certain fixed level of an overall safe outcome. One can specifically characterize such a HAV and traffic interactive strategy and use it as a safety performance indicator for the HAV. Under the proposed testing scheme, the HAV is evaluated under its full ODD with a reward function that represents a trade-off between safety and adversity in generating safety-critical events. The proposed methodology is demonstrated in simulation with various HAV designs under different operational design domains. 

\end{abstract}


\section{INTRODUCTION}

Highly automated vehicles (HAV) promise to significantly improve road safety and save lives. However, as the technology is evolving testing the safety performance and overall effectiveness is not straight-forward. There is a rich body of literature with many approaches to safety testing and various standards being applied throughout the industry and regulatory agencies but there aren’t universally accepted methods. Vehicles under test (VUT) are usually black-boxes to testers. Putting VUTs into real traffic and observing its performance may be a logical option but it is overly burdensome and unpractical, especially when considering continuous modifications in each version of the complex systems that constitute VUT. According to~\cite{nidhi2016driving}, 8.8 billion miles are needed to show its safety performance outperforms human drivers. The huge amount of time and resources required make such real-world testing approaches infeasible.  

Scenario-based testing has become commonplace to improve testing efficiency. As mentioned in~\cite{feng2021testing}, a test matrix is first determined based on real-world data like crash data and expert knowledge. With the test matrix, some may create a set of concrete scenarios with prescribed movements as in~\cite{van2017euro}, but this may make technology development biased to the predetermined concrete scenarios. Others may statistically generate scenarios based on the distribution in natural driving data (NDD) like~\cite{dingus2006100}. For this case, with the fast advancement of HAVs, the natural driving environment (NDE) changed significantly from consisting of only human drivers to a mix of human drivers and HAVs from various manufacturers. Handling the distribution change because of the NDE change can be a challenge. Both scenario generation methods suffer from their limited coverage of a VUT's operational design domain (ODD). 

Scenario generation from a test matrix parameter space covering the entire ODD of VUT is discussed in~\cite{weng2022formal}. With the increasing automation levels of HAVs, the parameter space of a test matrix grows exponentially. To address the "curse of dimensionality," deep reinforcement learning is used as mentioned in~\cite{feng2023dense} to generate testing scenarios. Moreover, testing a VUT under safety-critical events is essential. Due to the long tail and rare nature of safety-critical events in NDD, adversarial scenario generation is advocated in~\cite{capito2021modeled}~\cite{zhao2016accelerated} and~\cite{feng2021intelligent} to accelerate safety testing. Also, HAVs rely on their interactions with an environment for routing planning and decision-making. It is important to consider interactions in scenario generation as mentioned in~\cite{wang2022comprehensive}.

Once the scenarios are generated, one may test the VUT on the generated scenarios. Surrogate safety measures, including observed collision rate, time-to-collision (TTC), and others mentioned in~\cite{wang2021review} are then used to evaluate the testing results. The process of scenario generation and test result evaluation is commonly referred to as scenario-based testing.

\begin{figure}[!ht]
\centering
\includegraphics[width=0.45\textwidth]{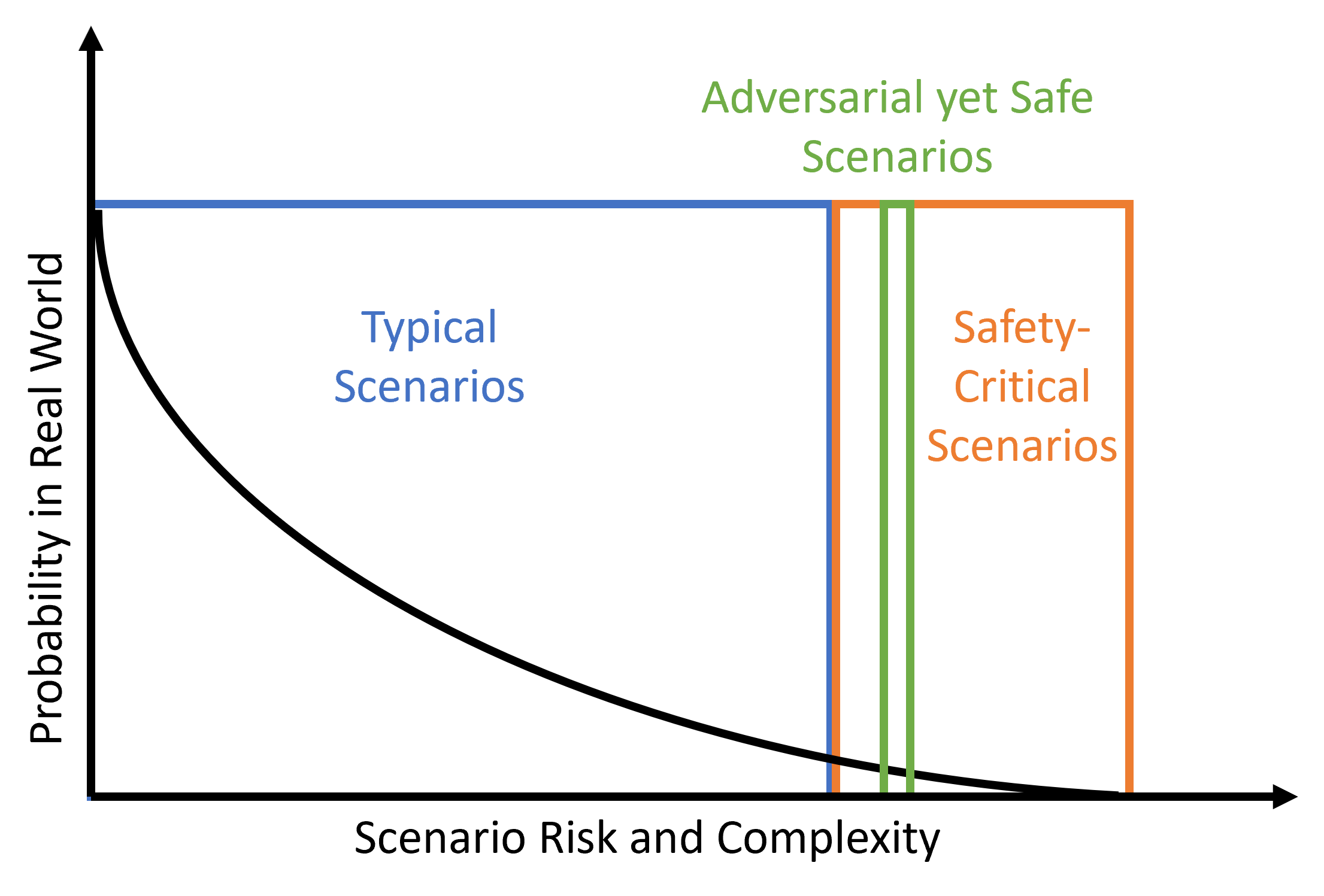}
\vspace{-3mm}
\caption{Typical scenarios that are less safety-related happen much more frequently in the real world compared with safety-critical scenarios. Safety performance can be better evaluated with safety-critical scenarios. We focus on a subset of safety-critical scenarios, called adversarial yet safe scenario (AYSS), to evaluate the safety performance of HAVs.}
\vspace{-7mm}
\label{fig:ayss}
\end{figure}

In this paper, we proposed a novel and informative scenario-based testing method for HAV safety performance characterization. The method, instead of testing all HAVs with a fixed set of scenarios, seeks to use principle other vehicle (POV) policies learned from interactions with different HAVs to characterize the safety performance of the HAVs. The POV policies are represented by neural networks. Scenarios are generated with their initial states sampled from a VUT's full ODD. To learn a POV policy, one or more POVs sharing the same POV policy are initialized around one or more VUTs sharing the same VUT policy to interact with the VUTs. The interactions collected are then used to optimize the POV policy based on a safety-related objective function. With the iterative executions of the interaction and optimization, a POV policy is selected as the safety performance characterization of the VUT policy based on the cost of the objective function. Different POV policies will be learned from the interactions with different VUT policies. To ensure fairness of the testing, all POV policies are optimized based on identical objective functions. Finally, with the intuition that a relatively safer VUT policy would allow the POV to exhibit a higher level of aggressiveness to achieve a certain fixed level of an overall safe outcome, the learned POV policies are used to generate adversarial yet safe scenarios (AYSS) as shown in Fig.~\ref{fig:ayss} and compare safety performance among corresponding VUT policies. We used two cases commonly seen in car following situations: one VUT following one POV in the same lane and one VUT handling cut-ins of one POV from an adjacent lane to show the effectiveness of our method. For simplicity, we assume other functionalities of VUTs, like lane centering, are perfect, and VUTs will not initiate lane changes during the test. Although we used car following for the study, our method is not limited to this domain and can be extended to systems with higher levels of automation and with more than one VUT and POV.  





Main contributions:
\begin{enumerate}
    \item We propose a new scenario definition $\{\mathcal{E}_{\pi}, \I, \X_{\pi'}\}$ which considers the interactions between VUTs and POVs. Observations of POVs are used to describe scenarios.
    \item AYSSs as shown in Fig.~\ref{fig:ayss} generated by POV policies learned from reinforcement learning are used to evaluate the safety performance of VUTs.
    \item The intuition that a relatively safe VUT policy would allow the POV to exhibit a higher level of aggressiveness to achieve a certain fixed level of an overall safe outcome is used to compare the safety performance of VUTs.
\end{enumerate}

This paper is organized as follows: we review related works in Section II. In Section III, we define the problem statement. The method is discussed in Section IV. Two case studies on car following functionality, including same lane following and cut-ins handling, are studied in Section V. Finally, we offer conclusions in Section VI.

\section{RELATED WORKS}

In~\cite{wang2022comprehensive}, the authors proposed a comprehensive HAV testing framework. It first generates a scenario library considering the interactions between VUTs and POVs using reinforcement learning. Then, scenarios in the library are adaptively selected considering the properties of a VUT for execution. Instead of selecting scenarios from a fixed scenario library, we generate AYSSs specific to VUTs.

As mentioned in~\cite{kuutti2020training}, two VUT policies are compared by training POVs to collide with VUTs. They use the first occurrence of collision in the training as a safety performance indicator. The VUT is considered relatively safer if its corresponding POV policy takes longer to learn how to cause a collision. However, the first occurrence of a collision may be affected by the optimization methoed. For example, when using stochastic gradient descent, if the differences between two VUTs are not significant, the safety performance of VUT may not be strongly correlated to the first occurrence of collision observed among multiple seeds. What's more, the scenario complexity and risk may be different when collision first occurs for different VUTs. Thus, the effort of learning to generate such scenarios may also be different. Thus, simply comparing the number of episodes needed to collide may not reflect the safety performance relation of VUTs for all cases. We propose to train POV policies on different VUTs with certain steps and use one criterion to pick a POV policy from all candidate POV policies obtained in training for a VUT. Then, the POV policies are evaluated on their observation space to analyze the safety performance of the corresponding VUT. 

Authors of~\cite{wang2022safety} argue that many existing safety performance testing methods consider a VUT policy as a white box, which is not true for many situations like third-party testing. They propose that to test a black box VUT policy, one can train a surrogate model with similar behaviors to the VUT. Then, Monte Carlo sampling can be used on selected key parameter spaces using the surrogate model to understand the safety performance of the VUT. As the complexity of VUT policy increases with the increasing automation level, finding an appropriate surrogate model for VUT behavior learning may be challenging. We use the same strategy to evaluate the learned model. We utilize Monte Carlo sampling on the observation space of learned POV policy instead of a surrogate model imitating the VUT.

\section{PROBLEM STATEMENT}

To characterize the safety performance of a VUT policy, we define an HAV testing environment containing $m$ VUTs and $n$ POVs where $m, n \in \mathbb{Z}_{\geq1}$. In the testing environment, VUTs can be described by a dynamic system, $f(\cdot)$, and share a driving policy, $\pi(\cdot)$. Consider the dynamic system
\begin{equation}\label{eq:dyn_sv}
    \s(t+1) = f(\s(t), \uu(t);\mathbf{w}(t))
\end{equation}
where VUT state $\s \in \Ss \subseteq \R^{s \times m}$, action $\uu \in \U \subseteq \R^{u \times m}$, and disturbance $\mathbf{w} \in \mathcal{W} \subseteq \R^{w}$, and the policy
\begin{equation}\label{eq:pi_sv}
    \uu(t)=\pi(\mathbf{x}(t))
\end{equation}
where VUT observation $\mathbf{x} \in \X \subseteq \R^{x \times m}$. Similarly, POVs can be described by a dynamic system, $f'(\cdot)$, and share a driving policy, $\pi'(\cdot)$, where
\begin{equation}\label{eq:dyn_pov}
    \s'(t+1) = f'(\s'(t), \uu'(t);\mathbf{w}(t))
\end{equation}
\begin{equation}\label{eq:pi_pov}
    \uu'(t)=\pi'(\mathbf{x}'(t))
\end{equation} 
with POV state $\s' \in \Ss' \subseteq \R^{s' \times n}$, action $\uu' \in \U' \subseteq \R^{u' \times n}$, and observation $\mathbf{x}' \in \X' \subseteq \R^{x' \times n}$. Testing will be executed in a structured environment with lane lines and speed limits defined. In this paper, we would like to investigate the VUT safety performance characterization problem in this testing environment and compare the different VUT policies, $\pi_1, \pi_2, \cdots$, with the safety performance characterizations. 

\section{METHOD}

\subsection{POV Policy Training}
The POV policy is represented by a deep neural network (DNN). Transitions of testing environment states can be modeled as a Markov Decision Process (MDP). Such MDP is defined by tuple $(\X', \U', \Pp, \Rr, \gamma)$, where $\X'$ is POV's observation, $\U'$ represents POV's action, $\Pp$ denotes state transition probability, $\Rr$ is the reward function, and $\gamma$ is the discount rate. At each time $t$, the POV observes state $x'_t$ and executes action $u'_t = \pi'(x'_t)$. The environment steps to next state according to $\Pp = p(x'_{t+1} | x'_t, u'_t)$. The POV is assigned a reward $r_{t} = \Rr(x'_{t})$. Now, a reinforcement learning problem is formulated to obtain an optimal POV policy $\pi^*$ by maximizing the expectation of discounted return, $\mathbb{E}[\sum_{k=0}^\infty\gamma^kr_{t+k}]$.

To obtain a POV policy that generates AYSSs, we design the reward function as 
\begin{equation}\label{eq:reward_function}
    \Rr(x'_t) = \alpha_A\Rr_A(x'_t) + \alpha_C\Rr_C(x'_t),
\end{equation}
where $\Rr_A$ is the adversarial reward that uses surrogate safety measures (SSM) like distance headway and two-dimensional TTC~\cite{guo2023modeling} to encourage safety critical states, $\Rr_C$ is the collision reward to penalize states related to a collision, and $\alpha_A$, $\alpha_C$ are their corresponding weights. The reward is a representation of the overall safety outcome.

\subsection{Scenario Generation}

\begin{definition}\label{def:scenario}
    \textbf{(Scenario)} To consider the interaction between VUTs and POVs, we define a driving scenario as a combination of three sets: $\z \in \mathcal{Z} = \{\mathcal{E}_{\pi}, \I, \X_{\pi'} \}$. Environment $\mathcal{E_{\pi}}$ including a VUT policy $\pi$ and testing environment like road shape, lane line, speed limit, etc., initial condition $\I$, and sequential observation $\X_{\pi'} = \{\Pp_{\pi'}(x', u')\} \subseteq \X^{'d}$ where $d$ is the number of steps of a scenario. 
\end{definition}

To generate scenarios, firstly, we fix a VUT policy $\pi$ and a testing environment. Then, initial conditions are sampled from the VUT's ODD. Lastly, POVs are placed with VUTs to obtain sequential observations for $d$ steps. 

\begin{figure}[!ht]
\centering
\includegraphics[width=0.45\textwidth]{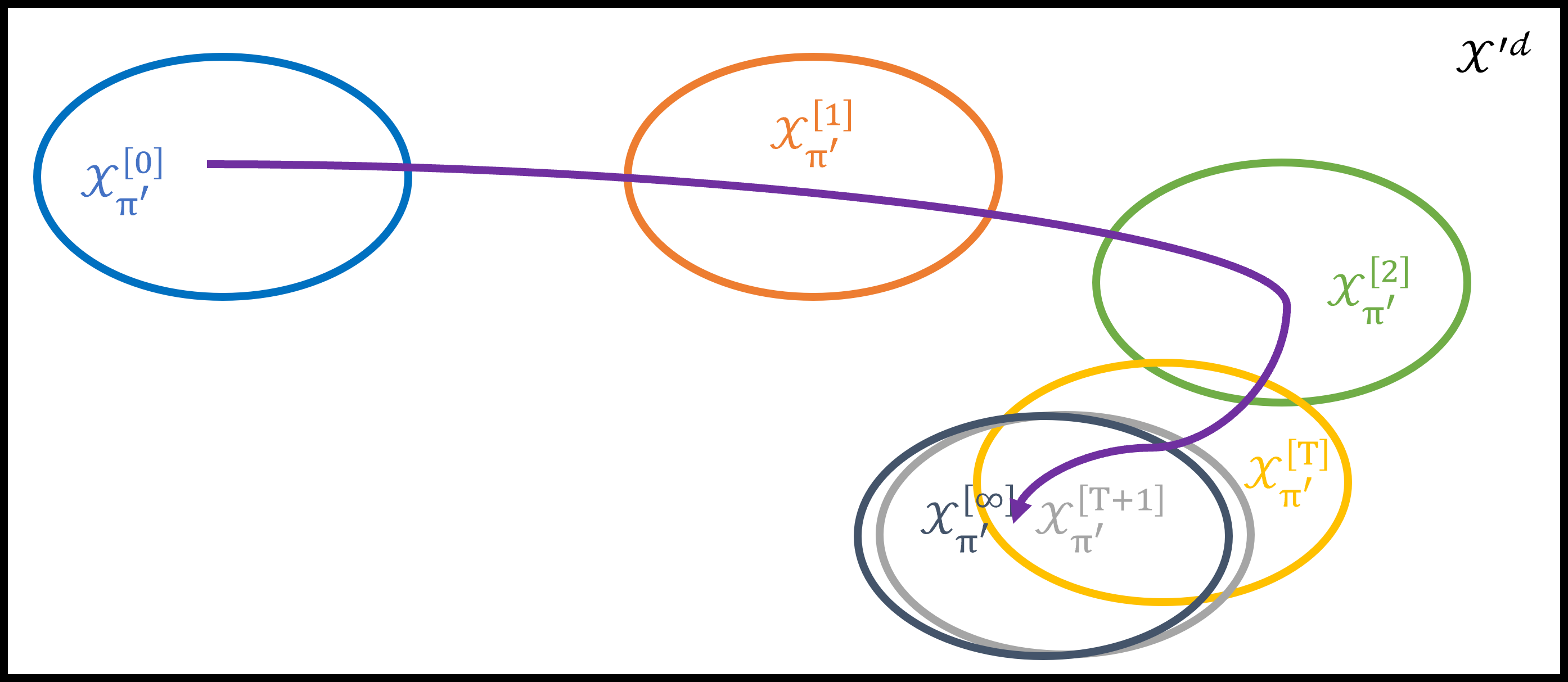}
\caption{Observations in generated scenarios change as the POV police updates.}
\vspace{-5mm}
\label{fig:pov_train}
\end{figure}

The data used to train the POV policy are the transitions $(x'_t, a'_t, r'_t, x'_{t+1})$, including the current observations, their associated actions, rewards, and next observations over testing steps. Transitions are collected in scenario generation and stored in a transition buffer. Iteratively, as shown in Fig.~\ref{fig:pov_train}, the policy is first updated by sample transitions in the buffer and then used to generate a new set of scenarios and add transitions included in new scenarios to the buffer. 

\subsection{Safety Performance Evaluation using AYSS}

\begin{definition}\label{def:SPCP}
    \textbf{(Safety Performance Characterization Policy (SPCP))} We defined the optimal POV policy $\pi^*$ obtained from training using generated scenarios as the SPCP of VUT policy $\pi$.
\end{definition}

To generate AYSSs with a $\pi^*$, we use initial states drawn from the observation space of $\pi^*$ with Monte Carlo sampling. It is unpractical to have Monte Carlo sampling with a black-box VUT policy. Thus, here, we use $d=1$ to generate scenarios as any states after the initial states cannot be obtained. As we focus on AYSSs, we do not want to generate typical scenarios or scenarios with unavoidable collisions. A scenario starting from a non-safety-critical state may take multiple steps to reach a safety-critical state and make it a safety-critical scenario. Since the scenario step is $1$, such scenarios should not be considered in the evaluation. Moreover, when a scenario causes unavoidable collisions, the obtained POV action can be random, as the outcome is the same no matter what action it takes. Therefore, such scenarios are not AYSSs and should not be considered.  

As all SPCPs are trained based on the same reward function that encourages adversarial yet safe scenarios, we argue that all SPCPs interact with their corresponding VUTs to achieve an identical level of overall safety outcome. With the intuition that a relatively safer VUT policy would allow the POV policy to exhibit a higher level of aggressiveness to achieve a certain fixed level of an overall safety outcome, we evaluate the aggressiveness levels using $\{\X_{\pi^*_1}\}$, $\{\X_{\pi^*_2}\}$, and $\{\X_{\pi^*_n}\}$  and aggressiveness indicators like brake magnitude and relative velocity projected on relative position~\cite{borrmann2015control} can be used for aggressiveness comparison.  

\subsection{Safety Performance Testing Framework}

\begin{figure}[!ht]
\centering
\vspace{-5mm}
\includegraphics[width=0.45\textwidth]{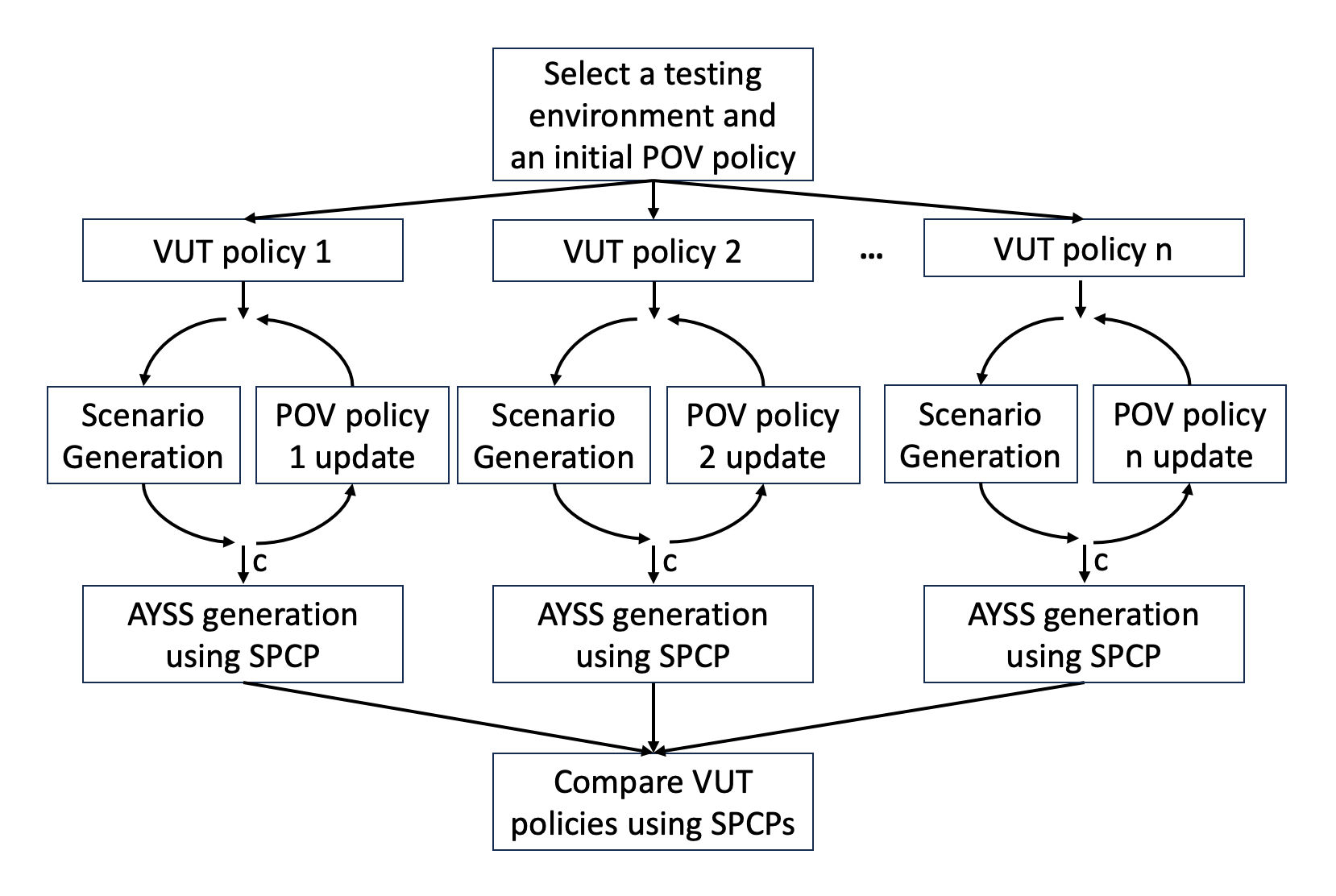 }
\vspace{-5mm}
\caption{Safety performance testing framework.}
\vspace{-5mm}
\label{fig:overall_process}
\end{figure}

As shown in Fig.~\ref{fig:overall_process}, to test the safety performance of a set of VUT policies $\{\pi_1, \pi_2, \cdots, \pi_n\}$, firstly, select a testing environment and an initial POV policy. For each VUT policy, a copy of the initial POV policy interacts with its respective VUT in the testing environment to generate a scenario $\{\mathcal{E}_{\pi_i}, \mathcal{I}, \mathcal{X}_{\pi_i^{\prime [t]}}\}$ where $d$ is the number of steps before updating POV policy $\pi_i^{\prime[t]}$. The POV policies of every update are the candidates of SPCP. After training is finished, one SPCP is selected for each VUT policy based on the same criteria $c$. Each SPCP is then used to generate the AYSSs. Finally, AYSSs of all VUT policies are used to evaluate their safety performance.


\section{CASE STUDY}

Double~\cite{van2016deep} dueling~\cite{wang2016dueling} DQN~\cite{mnih2013playing} with prioritized experience replay~\cite{schaul2015prioritized} implemented in Tianshou~\cite{weng2022tianshou} is used here to train the POV policy. We select discount rate $\gamma=0.8$ and learning rate $\text{lr}=10^{-5}$. The POV policy consists of 3 fully connected neural networks: feature extraction net, state value net, and advantage net. Each net contains two hidden layers. Each layer contains 256 neurons. ReLU is used as the activation function. SPCP candidates are stored and tested in 100 randomly generated scenarios every $10000$ steps in the testing environment. An average episode reward is recorded for each candidate. After training is finished, the candidate with the highest average episode reward is selected as the SPCP. 

Here, we choose the Intelligent Driver Model (IDM)~\cite{treiber2000congested} as the VUT policy. To have two different VUT policies, $\pi_1$ and $\pi_2$, as described in Table~\ref{table:vut_compare}, we assign minimum spacing $s_0=10$ m and $s_0=20$ m for the two VUT policies correspondingly with all other settings the same as in~\cite{leurent2018environment}. We argue that $\pi_2$ is relatively safer compared with $\pi_1$, as $\pi_2$ tends to keep a larger spacing with the front vehicle. 

We use Highway-Env~\cite{leurent2018environment} as the simulation environment to construct the testing environment with 5 Hz policy frequency and 5 seconds episode duration. VUT speed range is $[0, 40] \text{ m/s}$. VUT acceleration range is $[-6, 6] \text{ m/s}^2$. As implemented in~\cite{leurent2018environment}, with the knowledge of road structure (e.g., lane width, road network, etc.), the POV is controlled by acceleration and steering angle, which is calculated from target speed and current and target lane index. 

We studied two cases: one-lane car-following and two-lane cut-in. To focus on the car-following functionality modeled by IDM, we assume the VUT has perfect lane-centering functionality and will not seek to change lanes in any situation. In each case study, we first describe the experiment setup. Then, AYSSs aggressiveness comparison result is shown for VUT safety performance comparison. Lastly, SSM on AYSSs are studied to show that SPCPs generate AYSSs that achieve an identical level of safety outcome. 

\begin{table}
\begin{center}
\vspace{3mm}
\setlength\extrarowheight{1.5pt}
\begin{tabular}{ |c|c|c|c|c| } 
\hline
$\pi$ & $\mathbf{s}_0$ [m] & safety performance & expected $\pi^*$ behavior \\
\hline
$\pi_1$ & 10 & worse & more conservative \\ 
$\pi_2$ & 20 & better & more aggressive \\ 
\hline
\end{tabular}
\end{center}
\vspace{-2mm}
\caption{Two IDMs are used as two VUT polices, $\pi_1$ and $\pi_2$. All the IDM parameters of $\pi_1$ and $\pi_2$ are the same, except for minimum spacing $s_0$. $\pi_1$'s safety performance is worse as $\pi_1$ uses a shorter minimum spacing when making decisions than that of $\pi_2$. For SPCPs generating AYSSs to reach a certain overall safety outcome, $\pi^*_1$'s behavior is expected to be more conservative.}\label{table:vut_compare}
\vspace{-7mm}
\end{table}

\subsection{One-Lane Car-Following}

\begin{figure}[!ht]
\centering
\vspace{-5mm}
\includegraphics[width=0.45\textwidth]{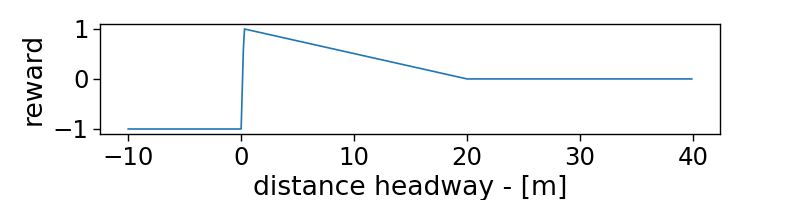}
\caption{Reward function used in one-lane car-following case study. The maximum reward is reached when distance headway $h=0.25$ m. No reward when $h>20$ m.}
\label{fig:case1_rew}
\vspace{-2mm}
\end{figure}

The testing environment is defined as a one-lane straight highway with one VUT with policy $\pi$ following one POV with policy $\pi'$. The VUT and POV are initialized at the center of the lanes with headings parallel to lane directions. The speed limit is 30 m/s. The POV observes $\{\mathbf{v}', \mathbf{v}, \mathbf{h}\}$ where $\mathbf{v}'$ is POV longitudinal speed, $\mathbf{v}$ is VUT longitudinal speed, and $\mathbf{h}$ is distance headway. The POV action space is defined as $\{u' = 0.0 + 0.5k\}, \forall k \in \{0, 1, \cdots, 20\}$ which sets the POV's target speed, $\mathbf{v}_{\text{target}} = \mathbf{u}'\cdot \mathbf{v}'$. The Reward function is shown in Fig.~\ref{fig:case1_rew}. To encourage adversarial behavior, the maximum reward occurs when $h=0.25$ m. The reward is gradually reduced and remains zero when $h>20$ m. Collision states are penalized with minimum reward. 

Three SPCPs are trained for each VUT policy with three seed values. We use multiple seeds to avoid getting a POV policy by coincidence. Then, we generate scenarios for each SPCP by conducting Monte Carlo sampling on its observation space. The observations in the generated scenarios are used to obtain their corresponding actions. Accelerations are calculated based on the actions. We remove typical scenarios by assuming distance headway is less than or equal to $5$ m. Scenarios with TTC $< 0.2$ s are identified as scenarios with unavoidable collisions. 

\begin{figure}[!ht]
\centering
\includegraphics[trim={0 2.3cm 0 1.7cm},clip,width=0.45\textwidth]{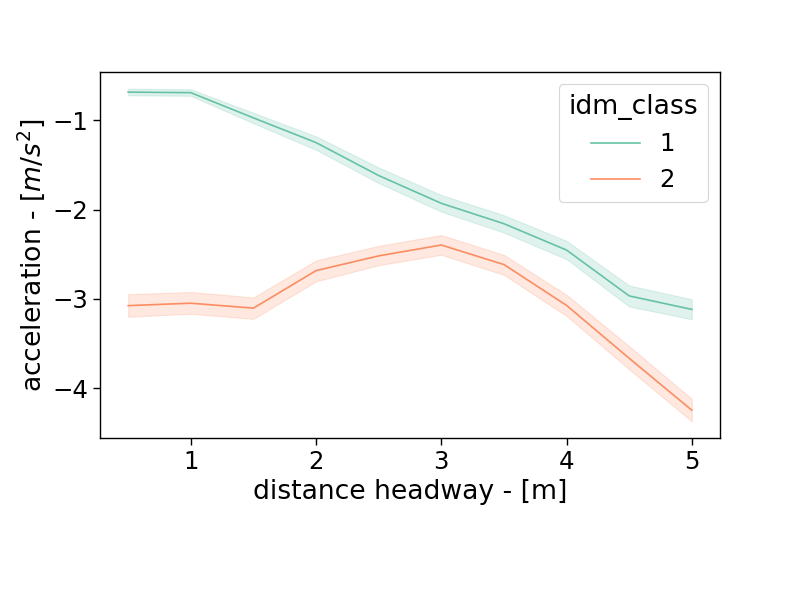}
\caption{Green and orange lines show the mean accelerations of the AYSSs generated by SPCPs trained using three seeds with $\pi_1$ and $\pi_2$ correspondingly over distance headways. The colored bands indicate their $95\%$ confidence intervals.}
\vspace{-8mm}
\label{fig:case1_result}
\end{figure}

By comparing sets of observations and actions obtained from generated scenarios, the result is shown in Fig.~\ref{fig:case1_result}. POVs trained with $\pi_1$ always brake less hard compared with those trained with $\pi_2$. This comparison result satisfies the expectation in Table~\ref{table:vut_compare}.

We also studied the overall safety outcomes of each SPCP. For each SPCP of the two VUT policies, we generated 200 scenarios using the same process as in the SPCP training. The overall safety outcomes of SPCPs between the two VUT policies are similar, as shown in Table~\ref{table:case1_overall_safety}. This shows that by using our method, the selected SPCPs can be used to generate AYSSs that achieve a similar level of safety outcome between VUTs with different safety performance.

\begin{table}
\begin{center}
\vspace{3mm}
\setlength\extrarowheight{1.5pt}
\begin{tabular}{ |c|c|c|c|c| } 
\hline
$\pi^*$ & $\pi$ & $\mathbf{c}$ [\%] & $\mathbf{\bar{r}}_{epi}$ & $\mathbf{\bar{h}}$ [m] \\
\hline
1 & $\pi_1$ & 0 & 19.48 & 1.65 \\ 
2 & $\pi_1$ & 0 & 19.45 & 1.82 \\ 
3 & $\pi_1$ & 0 & 19.43 & 1.71 \\ 
4 & $\pi_2$ & 0 & 19.30 & 2.42 \\ 
5 & $\pi_2$ & 0 & 19.31 & 2.62 \\ 
6 & $\pi_2$ & 0 & 19.29 & 3.76 \\ 
\hline
\end{tabular}
\end{center}
\vspace{-2mm}
\caption{Overall safety outcome study of one-lane car-following. Crash rate $\mathbf{c}$, mean episode reward $\mathbf{\bar{r}}_{epi}$, and mean distance headway $\mathbf{\bar{h}}$ at last step of each episode are calculated for each SPCP $\pi^*$ of the two VUT policies $\pi_1$ and $\pi_2$.}\label{table:case1_overall_safety}
\vspace{-3mm}
\end{table}

\subsection{Two-Lane Cut-In}
The testing environment is defined as a two-lane straight highway with one VUT with policy $\pi$ and one POV with policy $\pi'$. The POV is initialized on the left adjacent lane of the VUT. They are initialized on the center of their respective lanes with headings parallel to lane directions. The speed limit is 30 m/s. The POV observes $\{\mathbf{v}', \mathbf{v}, \mathbf{h}, \mathbf{l}\}$ where $\mathbf{v}'$ is the POV longitudinal speed, $\mathbf{v}$ is the VUT longitudinal speed, $\mathbf{h}$ is distance headway, and $\mathbf{l}$ is a binary value indicating whether the POV and the VUT are in the same lane. The POV action space is defined as the union of lane change action and target speeds, $\{\text{lane\_change}\} \cup \{u' = 0.0 + k\}, \forall k \in \{0, 1, \cdots, 40\}$. The reward function consists of three parts: same lane reward, different lane reward, and lane change reward. Firstly, it has no reward when the VUT and the POV are in different lanes, and the POV is not changing lanes. The rest are similar to the reward function shown in Fig.~\ref{fig:case1_rew}. While the POV is changing lanes to encourage adversarial behavior, it receives the maximum reward when $h=1$ m. The reward is gradually reduced and remains zero when $h>5$ m. Collision states are penalized with minimum reward. Lastly, the reward design is the same except for reaching maximum reward when $h=0.25$ m.


The POV policies are trained for each VUT policy three times with three seed values. We randomly generate $100$ test cases after every $1$ epoch to calculate a mean episode reward. The POV policy at the end of each epoch with the highest mean episode reward is selected as the SPCP. Then, we generate scenarios for each SPCP by conducting Monte Carlo sampling on its observation space. The observations in the generated scenarios are used to obtain their corresponding actions. Accelerations are calculated based on the actions. The VUT and POV are initialized on two adjacent lanes. As the scenario length is $1$, a scenario is safety-critical only if the action is "lane change." Also, lane changing is too risky when the POV and VUT are longitudinal overlapping along the road. Thus, scenarios with the POV rear bumper behind the VUT front bumper are not a part of the AYSSs. 

\begin{figure}[!ht]
\centering
\vspace{2mm}
\includegraphics[width=0.45\textwidth]{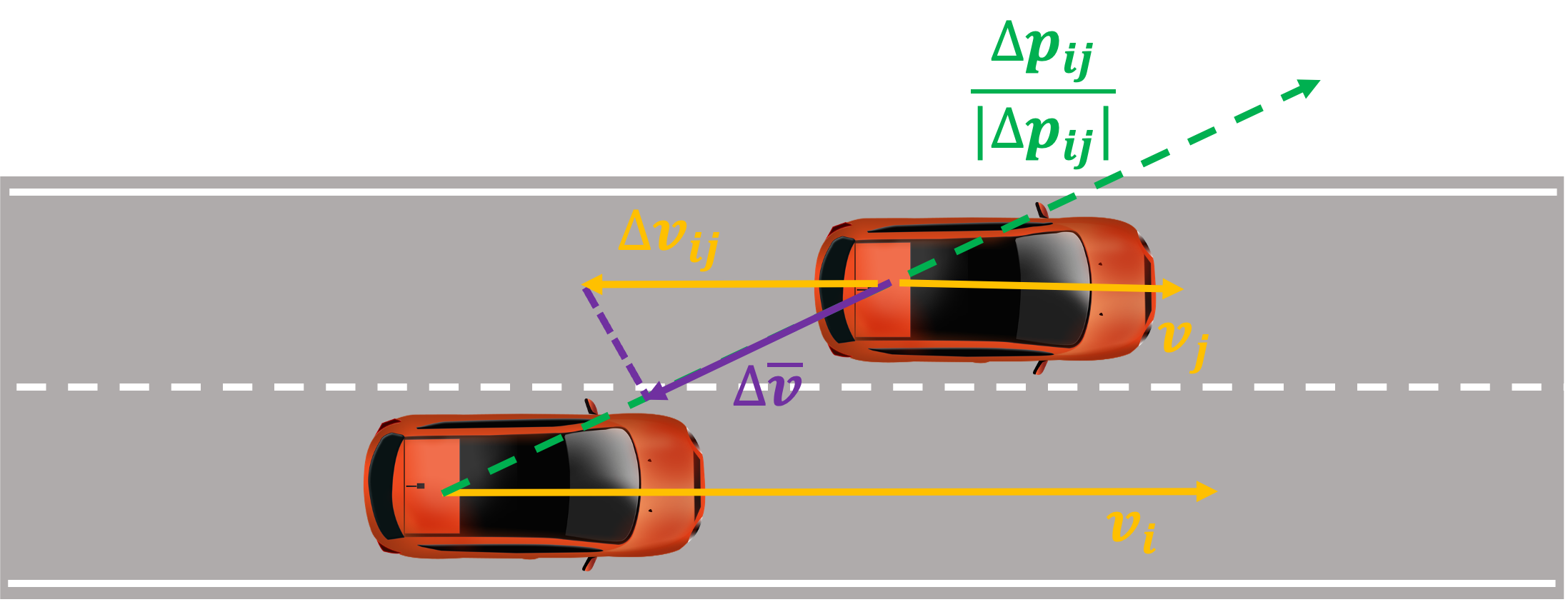}
\caption{Aggressiveness indicator $\Delta \bar{\mathbf{v}}$ described by the projection of relative velocity onto relative position vector. }
\vspace{-5mm}
\label{fig:delta_v_bar}
\end{figure}

We use $\Delta \bar{\mathbf{v}}$ proposed in~\cite{borrmann2015control} to describe the scenario aggressiveness here. Consider aggressiveness indicator 
\begin{equation}\label{eq:delta_v_bar}
    \Delta \bar{\mathbf{v}} = \frac{\Delta\mathbf{p}_{ij}}{||\Delta\mathbf{p}_{ij}||}\Delta\mathbf{v}_{ij},
\end{equation}
where $\Delta\mathbf{v}_{ij}$ is the relative velocity vector and $\Delta\mathbf{p}_{ij}$ is the relative position vector. $\Delta \bar{\mathbf{v}}$ is the projection of the relative velocity vector onto the relative position vector as shown in Fig.~\ref{fig:delta_v_bar}.~\cite{borrmann2015control} argues the relative velocity can be decomposed into $\Delta \bar{\mathbf{v}}$ and a vector perpendicular to the relative position vector. The latter vector acts to make one robot rotate around the other robot and hence does not affect the aggressiveness. Two robots are moving towards each other if $\Delta\mathbf{\bar{v}}$ is in the opposite direction of $\Delta\mathbf{p}_{ij}$. Hence, a system with smaller $\Delta \bar{\mathbf{v}}$ is more aggressive. As we need to consider the POV lateral movements in this case in addition to longitudinal movements, we use $\Delta \bar{\mathbf{v}}$ here to indicate scenario aggressiveness. 

\begin{figure}[!ht]
\centering
\includegraphics[trim={0 2.3cm 0 1.7cm},clip,width=0.45\textwidth]{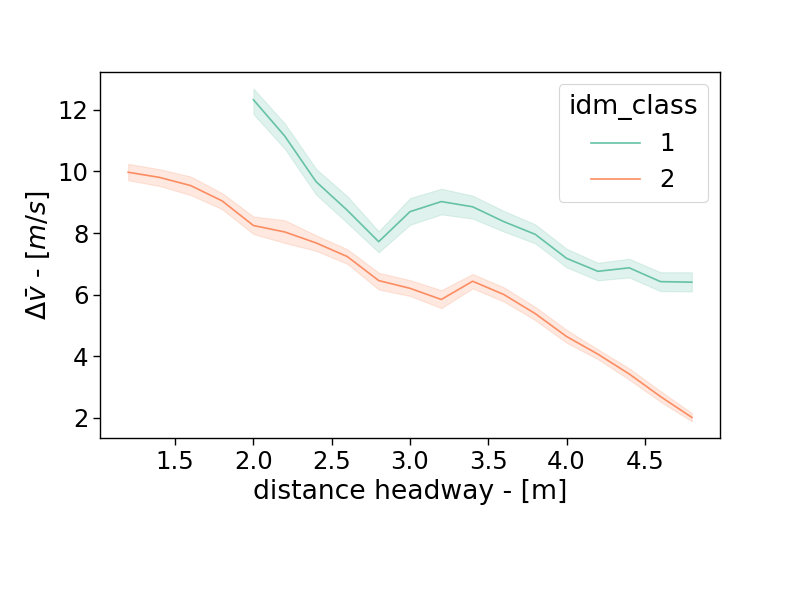}
\caption{Green and orange lines show $\Delta\mathbf{\bar{v}}$ of the AYSSs generated by SPCPs trained using three seeds with $\pi_1$ and $\pi_2$ correspondingly over distance headways. The colored bands indicate their $95\%$ confidence intervals.}
\vspace{-3mm}
\label{fig:case2_result}
\end{figure}

The generated scenarios between SPCPs of $\pi_1$ and $\pi_2$ are compared using $\Delta \bar{\mathbf{v}}$ over distance headway. As shown in Fig.~\ref{fig:case2_result}, with the same distance headways, $\Delta \bar{\mathbf{v}}$s of SPCPs of $\pi_1$ are greater than those of $\pi_2$. The result indicates that SPCPs of $\pi_2$ are more aggressive than SPCPs of $\pi_1$. Consequently, $\pi_2$ has better safety performance compared with $\pi_1$. This conclusion matches the IDM model analysis shown in Table~\ref{table:vut_compare}.

Similar to the one-lane car-following case, we studied the overall safety outcome of scenarios generated by SPCPs. In addition, we calculated the lane change rate. The result is shown in Table~\ref{table:case2_overall_safety}. $\bar{\mathbf{r}}_{epi}$ and $\bar{\mathbf{h}}$ are calculated using scenarios with lane changes and without collisions. This reflects the same conclusion drawn from one-lane car-following case that the selected SPCPs can be used to generate AYSSs that achieve a similar level of safety outcome between VUTs with different safety performance.

\begin{table}
\begin{center}
\setlength\extrarowheight{1.5pt}
\begin{tabular}{ |c|c|c|c|c|c| } 
\hline
$\pi^*$ & $\pi$ & $\mathbf{c}$ [\%] & lc [\%] & $\mathbf{\bar{r}}_{epi}$ & $\mathbf{\bar{h}}$ [m] \\
\hline
1 & $\pi_1$ & 1.0 & 99 & 20.99 & 0.36 \\ 
2 & $\pi_1$ & 2.0 & 99 & 20.74 & 0.36 \\ 
3 & $\pi_1$ & 0.5 & 98 & 20.78 & 0.36 \\ 
4 & $\pi_2$ & 3.5 & 100 & 20.41 & 0.41 \\ 
5 & $\pi_2$ & 1.0 & 100 & 20.93 & 0.38 \\ 
6 & $\pi_2$ & 1.0 & 100 & 21.14 & 0.35 \\ 
\hline
\end{tabular}
\end{center}
\vspace{-2mm}
\caption{Overall safety outcome study of two-lane cut-in. Crash rate $\mathbf{c}$, lane change rate lc, mean episode reward $\mathbf{\bar{r}}_{epi}$, and mean distance headway at last step of each episode $\mathbf{\bar{h}}$ are calculated for each SPCP $\pi^*$ of the two VUT policies $\pi_1$ and $\pi_2$.}\label{table:case2_overall_safety}
\end{table}


\section{CONCLUSIONS}

In this paper, we proposed a black-box HAV safety performance testing framework and demonstrated its application in simulation. The framework follows the intuition that a relatively safe VUT policy would allow the POV to
exhibit a higher level of aggressiveness to achieve a certain fixed level of an overall safe outcome. We first train POVs by interacting with VUTs to reach a certain overall safety outcome. One SPCP is selected for each VUT. SPCPs of all VUTs are selected based on the same criteria. Then, AYSSs are generated for each SPCP. Lastly, we compare the safety performance of VUTs by evaluating the AYSSs. We demonstrate the effectiveness of the testing framework by studying two cases: one-lane car-following and two-lane cut-in. In each case study, we gave examples of the testing environment, the POV training setup, the AYSSs generation, and their aggressiveness evaluation, and studied the SPCPs' overall safety outcomes. Although we studied VUTs with level-1 HAV features here, the testing framework can be extended to higher levels of automation. The testing framework is not limited to a single VUT. More POVs can also be accommodated in this framework. Lastly, accelerating the POV training can reduce the test burden with less physical VUT's involvement.   








\bibliographystyle{unsrt}
\bibliography{root}

\end{document}